\documentclass[letterpaper, 10 pt, conference]{ieeeconf}  

\IEEEoverridecommandlockouts                              

\newcommand{\T}{{\sf T}}

\newcommand{\pp}[2]{\frac{\partial #1}{\partial #2}}

\usepackage{subcaption}
\usepackage{floatrow}
\usepackage{graphicx} 
\usepackage{comment}
\usepackage{cite}
\usepackage{algorithm}
\usepackage{algorithmic}
\usepackage{amsmath} 
\usepackage{amssymb}
\usepackage{bm}
\usepackage{balance}
\usepackage{hyperref}
\hypersetup{
    colorlinks=true,
    linkcolor=black,
    filecolor=blue,      
    urlcolor=black,
    pdftitle={BuzzRacer},
    pdfpagemode=FullScreen,
    }

\usepackage{xcolor}
\floatstyle{plaintop}
\restylefloat{table}
\usepackage[tableposition=top]{caption}

\title{\LARGE \bf
Residual Descent Differential Dynamic Game (RD3G) -- A Fast Newton Solver for Constrained General Sum Games 
}

\author{Zhiyuan Zhang$^{1}$ and Panagiotis Tsiotras$^{2}$
\thanks{This work is sponsored by ONR awards N00014-23-2308 and N00014-23-2353 and NSF award IIS-2008686}
\thanks{$^{1}$ School of Aerospace Engineering, Institute for Robotics and Intelligent Machines, Georgia Institute of Technology, Atlanta, GA 30332, USA, Email:
        {\tt\small zzhang615@gatech.edu}}%
\thanks{$^{2}$ School of Aerospace Engineering, Institute for Robotics and Intelligent Machines, Georgia Institute of Technology, Atlanta, GA 30332, USA, Email:
        {\tt\small tsiotras@gatech.edu}}%
}

\begin{document}
\maketitle
\thispagestyle{empty}
\pagestyle{empty}

\begin{abstract}
We present Residual Descent Differential Dynamic Game (RD3G), a Newton-based solver for constrained multi-agent game-control problems.
The proposed solver seeks a local Nash equilibrium for problems where agents are coupled through their rewards and state constraints.
We compare the proposed method against competing state-of-the-art techniques and showcase the computational benefits of the RD3G algorithm on several example problems.
\end{abstract}

\section{Introduction}

Differential dynamic games (DDG) have a wide range of applications in robotics.
Many complex problems in multi-agent motion planning in particular, such as drone racing \cite{gt_1}, autonomous driving \cite{gt_2}, satellite control \cite{satellite}, etc.,
can be modeled as DDGs, necessitating effective numerical methods to find their solutions. 

Game-theoretic control seeks to find an equilibrium to a control problem modeled as a DDG.
In game-theoretic problems, 
it is common to assume decoupled dynamics, that is, 
each agent's control affects only that agent's state.
The agents interact with each other through mutual state constraints and through the game objective.
Some commonly used solution concepts for game-theoretic control include Nash equilibrium~\cite{nash}, Stackleberg equilibrium~\cite{Stackleberg}, and Correlated equilibrium~\cite{correlated}.
While the most suitable solution concept depends on the particular problem, Nash equilibrium remains the most studied and widely used equilibrium concept because it does not assume the order of action taken between agents (e.g., Stackleberg equilibrium), 
nor does it require some form of external coordination (e.g., Correlated equilibrium). 
It is therefore more suitable for uncooperative interactions, which is common in many robotic control problems.

\subsection{Related Work}

Various approaches have been proposed to find the Nash equilibrium of a DDG.
A common approach is to discretize the action space and convert the differential dynamic game into a bimatrix game~\cite{bimatrix}.
Since the action space can be large, good heuristics are needed to choose a manageable set of candidate trajectories.
A key advantage of this approach is that the trajectories may be computed in parallel  on a Graphics Processing Unit (GPU),
and then the bimatrix game can be solved using existing methods. This method works well in practice but relies heavily on heuristics.
However, this method does not find the equilibrium of the original game, which has a continuous action space.

The Iterated Best Response (IBR)~\cite{ibr} iteratively solves the single-agent optimal control problem for each agent in the game while holding the policy of the other agents static. 
There is no guarantee of convergence, but if IBR converges for a particular problem, it does converge to the Nash equilibrium~\cite{ibr}.  
IBR, however,  suffers in terms of convergence speed since agent interaction has to manifest through mutual constraints in multiple iterations. 

Sensitivity-Enhanced IBR (SE-IBR)~\cite{seibr} is a recent improvement over IBR with a modified cost function to encourage the agent to move in the direction of the agent collision constraint, therefore discovering potential interactions, thus accelerating convergence.
Still, SE-IBR involves solving multiple constrained optimal control problems at each time step, when used in a receding horizon fashion and depends on a high-performance solver to work in realtime.

While the equilibrium for a DDG with continuous action space and nonlinear dynamics is generally hard to find, a Linear Quadratic Game (LQ Game)  admits an analytical solution~\cite{gt_textbook}. 
Therefore, 
several methods approximate a nonlinear DDG with a multi-stage LQ Game and use Differential Dynamic Programming (DDP) to find an open- or closed-loop equilibrium control policy. 
Minimax DDP~\cite{minimax_ddp} treats each stage as a minimax game, while iLQGame \cite{ilqgame} and OPTGame3 \cite{optgame} extend Minimax DDP to general sum LQ Games. iLQGame, in particular, has been used successfully on motion control problems like autonomous car racing \cite{ilqgame_racing1}, \cite{ilqgame_racing2}.

The approaches discussed above are homotopy-type methods that approximate the original game using a series of easier-to-solve problems whose solutions converge to the equilibrium of the original game.
However, it is also possible to directly solve for the necessary conditions for a Nash equilibrium\cite{algames,dldu_1,dldu_2}. 
Earlier methods such as \cite{dldu_1,dldu_2} use a single shooting approach and choose a descent step in the direction of $\textrm{d}L/\textrm{d}u$, where $L$ is the Lagrangian and $u$ is the action variable.
ALGame~\cite{algames} is a multiple-shooting Newton method that finds the root of the KKT necessary conditions equations,
and uses a dual ascent update step to manage the inequality constraints.
For some robotic problems, ALGame is reportedly faster than iLQGame~\cite{algames}.

\subsection{Contributions}

In this work, we propose a novel Newton-based method to solve numerically game-theoretic control problems. 
In most Newton methods, solving for the descent direction is the most time-consuming part of the algorithm~\cite{convex_opt}. 
The proposed RD3G algorithm reduces the scale of the linear problem associated with finding the descent direction by partitioning the collision constraints and by removing the dual variables associated with the inactive constraints from the problem, 
and then uses an interior point method along with a barrier function to maintain solution feasibility for the inactive constraints. 
This technique decreases the time spent in each iteration step by a large margin. 
In addition, we use a multiple shooting technique and perform gradient descent on the states and the controls simultaneously to help convergence.

Many existing methods for game-theoretic control are single-shooting 
methods~\cite{dldu_1,dldu_2,ilqgame,seibr}, which treat the state trajectory purely as derived variables of the control variables. 
This technique eliminates the need for dynamic constraints and results in simpler formulations. 
However, the derivatives of the state variables with respect to the control variable suffer from the diminishing gradient problem~\cite{vanishing}.
Moreover, an auto-gradient library 
is often needed to provide gradients due to the complex structure of the compute graph. 

The proposed method uses indirect multiple shooting and takes gradient descent steps to solve the necessary conditions for a Nash equilibrium, 
which improves numerical stability and convergence, particularly for longer time horizons or stiff problems. 
It also mitigates issues related to the sensitivity of the solution to initial conditions, as it avoids the accumulation of numerical errors over the entire time horizon~\cite{multiple_shooting}.

The remainder of the paper is organized as follows: Section~\ref{sec:formulation} formulates the game, and Section~\ref{sec:methods} describes the proposed solution method in detail. Section~\ref{sec:simulations} shows several example problems, benchmark results, 
and physical experiments, followed by a short discussion in Section~\ref{sec:discussion}. 

\section{Problem Formulation}\label{sec:formulation}

Let $x_k^i\in \mathbb{R}^n$ and $ u_k^i\in \mathbb{R}^m$ denote the state and control vectors, respectively, for agent $i$ at timestep $k$. 
To keep the notation simple, we assume that all agents have the same dimension for the state and control. However, the approach can be easily extended to the case of heterogeneous agents. 
We use $ [a..b] $ to denote all the integers between $a$ and $b$, including $a$ and $b$.
We use parentheses to denote the concatenation of vector variables, i.e., $ (a_1, a_2, \dots, a_n)  = [a_1^\T, a_2^\T, \dots, a_n^\T]^\T$. 
When we omit the superscript of a variable, it holds for all agents. For example, $x_k = (x_k^1, x_k^2, \dots, x_k^N)$ is the concatenated states of all agents at time step $k$.
The same notation applies for omitted subscripts, denoting every time step,
that is, $x^i = (x_1^i, x_2^i, \dots, x_T^i)$ denotes the state trajectory for agent $i$ at all time steps. 
The game state $x$ is the concatenated state of all agents, i.e., $x = (x^1, x^2, \dots, x^N)$. 
Finally, $u^{-i}$ denotes the concatenated control of all agents except agent $i$. 
Let $N$ be the number of agents and $T$ be the horizon length.

We are interested in finding the generalized Nash equilibrium to the following game:

\begin{subequations}\label{eq:2}
\label{eq:formulation}
\begin{align}
    & \underset{u^i}{\min}\, J^i (x, u^i), \quad \forall i \in [1..N] \label{eq:2a},\\
     &x^i_{k+1} = f^i(x_k^i, u_k^i)  , \quad \forall k \in [0..T-1], \forall i \in [1..N]\label{eq:2b},\\
    &h(x^i_k, x^j_k) \leq 0, \quad \forall i\neq j, i,j\in[1..N], \forall k\in[1..T]\label{eq:2c},
\end{align}
where,
\begin{align}
    &J^i(x,u^i) = \sum_{k=0}^{T-1} J_k^i(x_k, u_k^i) + \phi^i(x_T),
\end{align}
\end{subequations}
denotes the total cost for agent $i$, which is the sum of step cost $J_k^i(x_k, u_k^i)$ and the terminal cost $\phi^i(x_T)$.
In (\ref{eq:2b}) $f^i(x^i_k,u^i_k)$ denotes the dynamics for agent $i$, and in (\ref{eq:2c}) $h(x^i_k, x^j_k)$ is the agent-to-agent interaction constraint,
which, without loss of generality, is assumed to be the same for all agents.

We seek the open-loop Nash equilibrium to the Generalized Nash Equilibrium Problem (GNEP) described in (\ref{eq:2}), that is, 
a set of control actions for all agents such that no agent can decrease its cost objective by unilaterally changing its control action.  
We denote the Nash equilibrium control as $u^*$.

\section{Solution Approach}
\label{sec:methods}

The proposed method contains four steps. First, partition the inequality state constraints in (\ref{eq:2c}).
Second, find the descent direction of (\ref{eq:formulation}).
Finally, determine the step size with a line search. 

Our algorithm starts with an initial guess of the control and state trajectory. 
This initial guess does not have to satisfy the state constraints in (\ref{eq:2c}) but it should satisfy the dynamics constraints in (\ref{eq:2b}). 

\subsection{Constraint Partition}
The number of inequality constraints $h(x^i_k, x^j_k)$ grows quadratically with the number of agents and linearly with the time horizon. 
Each constraint adds another dual variable, increasing the size of the decision variables, thus negatively impacting performance.
To reduce the number of constraints, 
we separate the inequality constraints into two types: those currently not satisfied,
which we need to rectify,
and those already satisfied, which we maintain using a barrier function. 
 
Denote the index set of active and currently violated constraints as $H^+_{i,k} = \{j | h(x_k^i,x_k^j) \geq 0,  i,j \in [1..N], j\neq i \}$, and the set of inactive constraints as $H^-_{i,k} = \{j | h(x_k^i,x_k^j) < 0, i,j \in[1..N] , j\neq i\}$.
We define the step Lagrangian for agent $i$ as 
\begin{align}\label{eq:partition}
    L^i_k &= J^i_k(x_k, u^i_k) + \lambda^{i,\T}_k [f^i(x_k^i, u_k^i) - x_{k+1}^i] \nonumber \\
    &+ \sum_{j\in H^+_{i,k}} \mu^{i,j}_k h(x_k^i, x_k^j) + \sum_{j\in H^-_{i,k}} \mathrm{B}( h(x_k^i, x_k^j) ),
\end{align}
where $\mathrm{B}(h) = -\rho \log(-h)$ is the one-sided barrier function, and $\rho > 0$ is a homotopy parameter in (\ref{eq:partition}), and $i\in[1..N], k \in [0..T-1]$. In~(\ref{eq:partition})  
$\lambda^i_k$ is the Lagrange multiplier for the $i$\textsuperscript{th} agent dynamics constraint.
Define the variable $\lambda \in \mathbb{R}^{TN}$ as the concatenation of all $\lambda^i_k$ for all $i$ and $k$, i.e., $\lambda = (\lambda_0, \lambda_1, \dots, \lambda_{T-1})$, where $\lambda_k = (\lambda^1_k, \lambda^2_k, \dots, \lambda^N_k)$.
Similarly, $\mu^{i,j}_k$ is the Lagrange multiplier for the inequality state constraint, and let 
$ \mu \in \mathbb{R}^{TN^2} $ be the concatenation of all $\mu^{i,j}_k$, defined by 
$\mu = (\mu_1, \mu_2, \dots, \mu_T)$, where $\mu_k = (\mu_k^{1,1}, \mu_k^{2,1}, \dots, \mu_k^{N,1}, \mu_k^{1,2}, \mu_k^{2,2}, \dots, \mu_k^{N,2}, \dots, \mu_k^{N,N})$.

The Lagrangian for agent $i$ is
\begin{align}
    \mathcal{L}^i =& \sum_{k=1}^{T-1} L_k^i + J_0^i(x_0^i, u_0^i) + \lambda_0^{i,\T}[f^i(x_0^i,u_0^i) - x_1^i] + \phi^i(x_T^i) \nonumber \\
    &+ \sum_{j\in H^+_{i,T}} \mu^{i,j}_T h(x_T^i, x_T^j) + \sum_{j\in H^-_{i,T}} \mathrm{B}( h(x_T^i, x_T^j) ),
\end{align}
where, for notational simplicity, we retain the term $J_0^i(x_0, u_0^i)$ to penalize the control at the first time step for notational simplicity, even though $x_0$ is not part of the decision variable. 

\subsection{Residual Descent}
The necessary conditions for the generalized Nash equilibrium of the converted problem for agent 
$i$ are~\cite{gnep}
\begin{subequations}
\begin{align}
\pp{\mathcal{L}^i}{x^i} &= 0, \\
\pp{\mathcal{L}^i}{u^i} &= 0,\\
f^i(x_k^i,u_k^i) - x^i_{k+1} &= 0, \quad \forall k\in[0..T-1],\\
\mu^{i,j}_k > 0, \, h(x^i_k, x^j_k)&=0, \quad \forall j\in H^+_{i,k}, \forall k\in[1..T].
\end{align}
\end{subequations}
Next, let $F^i = ( f^i(x_0^i, u_0^i) - x^i_{1},f^i(x_1^i, u_1^i) - x^i_{2},\dots,f^i(x_{T-1}^i, u_{T-1}^i) - x^i_T) $, and $h^{i,+} = ( h_1, h_2, \dots, h_T )$, where $h_k = (h(x^i_k, x^{j_1}_k), h(x^i_k, x^{j_2}_k),\dots,h(x^i_k, x^{j_M}_k) )$, where $j_1,\dots,j_M$ are the enumeration of elements in $H^+_{i,k}$, with $M = |H^+_{i,k}|$.
Define the residual of agent $i$ as
\begin{equation}
    r^i(x,u,\lambda, \mu) = \left(\pp{\mathcal{L}^i}{x^i},\pp{\mathcal{L}^i}{u^i},F^i, h^{i,+}\right),
\end{equation}
and let the concatenated residual for all agents be $r = (r^1, r^2, \dots, r^N)$.
Denote the concatenated decision variables as $y=(x,u,\lambda,\mu^+)$, 
where $\mu^+$ stands for all entries of $\mu$ that are associated with an active or violated inequality state constraint.
The objective of the Newton method is to find a solution to the set of equations  $r^i(x,u,\lambda, \mu) =0$ for all $i\in[1..N]$ by minimizing the norm $\|r(x,u,\lambda, \mu)\|^2$.

To this end, let $y_\ell$ be the current value for $y$, where $\ell$ denotes the current iteration step.
We find the descent direction 
for the primal-dual pair $y$ by solving for $\Delta y = (\Delta x,\Delta u,\Delta \lambda,\Delta \mu^+)$ from the first-order Taylor expansion for the combined residual, as in the following equation:
\begin{equation}
\label{eq:taylor}
    r(y_\ell) + \nabla r(y_\ell) \Delta y = 0,
\end{equation}
where $\nabla r(y_\ell)$ is the Jacobian matrix of $r$ evaluated at $y_\ell$. Note that we only retain the entries in $\mu$ associated with the active or currently violated constraints to reduce the dimension of $\Delta y$. 
The solution to (\ref{eq:taylor}) is the change to $y_\ell$ that drives the residual $r(y_\ell + \Delta y)$ to zero, given the first order approximation, hence a descent direction.

\subsection{Backtracking Line Search}
We perform a backtracking line search in the descent direction to ensure $\|r(y_\ell+\Delta y)\|$ decreases with the chosen step size, and that the descent step does not increase the state constraint residual. 
It is important to check that a previously satisfied state constraint does not get violated when we take a descent step. 
We define the sum of all violated state constraint residuals as 
$\mathbf{h}(x) = \sum_{\{i,j,k | h(x^i_k, x^j_k)> 0\}} h(x^i_k, x^j_k)$. Algorithm~\ref{alg:line_search} describes the line search in detail.

\begin{algorithm}
\caption{ Backtracking line search }
\begin{algorithmic}
    \STATE  Define constants $\beta \in (0,1), \alpha \in (0,1) $, set variable $t=1$\\
    \STATE Define $(x, u, \lambda, \mu) \leftarrow y_\ell, \quad (\Delta{x}, \Delta{u}, \Delta{\lambda}, \Delta{\mu}) \leftarrow \Delta{y}$ \\
    \STATE $y_t = [x + t\Delta x, u + t\Delta u, \lambda + t\Delta \lambda, \mu + t\Delta \mu ]$ \\
    \WHILE {$\|r(y_t)\| > (1-\alpha t) \|r(y_\ell)\|$ or $\mathbf{h}(x + t\Delta x) > \mathbf{h}(x)$}
    \STATE $t \leftarrow \beta t$
    \ENDWHILE
\end{algorithmic}
\label{alg:line_search}
\end{algorithm}

We update the homotopy parameter $\rho$ with the strategy proposed in~\cite{homotopy}, which is used in the popular solver IPOPT \cite{ipopt}.
The update rule is:

\begin{equation}
\label{eq:homotopy}
\rho_{\ell+1} = \max \left\{ \frac{\epsilon_{\mathrm{tol}}}{10}, \min \left\{ \kappa \rho_\ell, \rho_\ell^\beta \right\} \right\},
\end{equation}
where $\epsilon_{\mathrm{tol}}$ is the desired tolerance, 
 and $\kappa, \beta$ are tuning parameters.

With this strategy, the homotopy parameter eventually decreases at a superlinear rate but does not get too small to cause numerical stability problems.
The choice of $\rho$ in (\ref{eq:homotopy}) is proven to give rise to superlinear convergence under standard second-order sufficient conditions~\cite{homotopy}.
Algorithm~\ref{alg:full} summarizes the proposed algorithm.

\begin{algorithm}
\caption{Residual Descent Differential Dynamic Game}
\begin{algorithmic}
\STATE Initialize $u, \lambda=0, \mu=0$, compute initial $x$ from $u$ 
\REPEAT
    \STATE Evaluate $h(x^i_k,x^j_k)$, build sets $H^+_{i,k}, H^-_{i,k}$ for all agents
    \STATE Calculate gradient of residual $\nabla r(y_\ell)$
    \STATE Solve for descent direction $\Delta y$ from (\ref{eq:taylor})
    \STATE Perform line search, find step size $t$ using Algorithm~1
    \STATE Update $y_{\ell+1} \leftarrow y_\ell + t\Delta y$
    \STATE Update homotopy parameter $\rho$ with (\ref{eq:homotopy})
\UNTIL{$\|r(y_\ell)\| < \epsilon_{\mathrm{tol}}$}
\RETURN $y_\ell$
\end{algorithmic}
\label{alg:full}
\end{algorithm}
\section{Experiments}
\label{sec:simulations}
\subsection{Implementation}
To evaluate the algorithm, the solver was implemented using Python and C++, with Eigen3~\cite{eigen3} for matrix manipulation in C++.  
The Least Squares Conjugate Gradient (LSCG)~\cite{lscg} was used as the main solver for the descent direction.
In case LSCG fails to converge, 
the implementation falls back to SparseQR~\cite{sparseQR}, which takes longer to converge but is more robust.

All experiments were performed on a desktop computer with an Intel i7-7700k 4.2Hz CPU running Ubuntu 22.04.

\subsection{Numerical Examples}

\subsubsection{Car Merging}\label{sec:car_merge}

The traffic merging problem is a classic example of a differential game~\cite{merge}. 
The game starts with several cars traveling in two adjacent lanes, each with a different longitudinal speed. 
The cars in the right lane attempt to merge into the left lane. 
Each car follows the kinematic bicycle model~\cite{kinematic} with $[x, y, v, \theta]$ as the states, 
corresponding to the longitudinal position, lateral position, speed, and heading, respectively. 
The control for each car $u=[\delta_T, \delta_S]$ is the acceleration and the steering angle, respectively. 

\begin{figure}[h]
  \centering
  \begin{subfigure}[b]{0.20\textwidth}
  \includegraphics[trim=290px 45px 280px 170px,clip,width=0.95\textwidth]{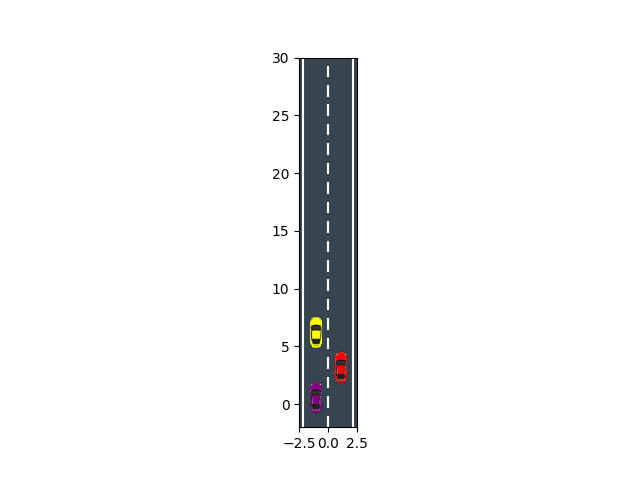}
  \caption{}
  \end{subfigure}
  \begin{subfigure}[b]{0.20\textwidth}
  \includegraphics[trim=290px 45px 280px 170px,clip,width=0.95\textwidth]{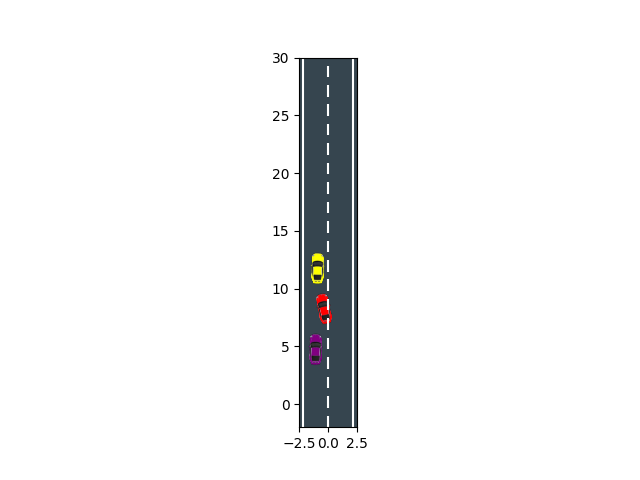}
  \caption{}
  \end{subfigure}
  \begin{subfigure}[b]{0.20\textwidth}
  \includegraphics[trim=290px 45px 280px 170px,clip,width=0.95\textwidth]{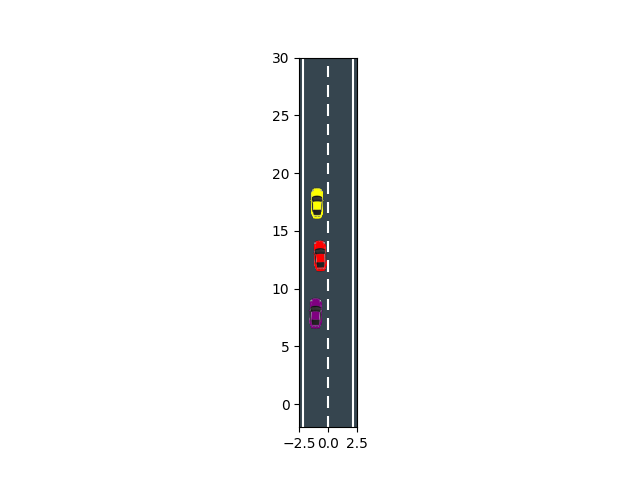}
  \caption{}
  \end{subfigure}
  \begin{subfigure}[b]{0.20\textwidth}
  \includegraphics[trim=210px 45px 200px 110px,clip,width=0.95\textwidth]{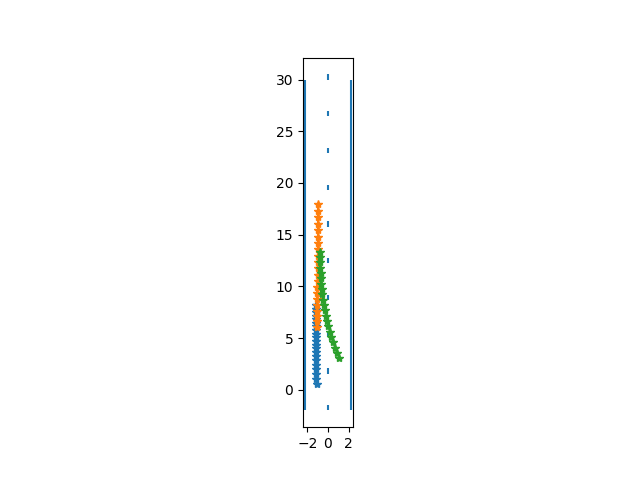}
  \caption{}
  \end{subfigure}
  \caption{Merging example with three cars,
  (a)-(c): Snapshots during merge, 
  (d) Trajectory plot.}
  \label{fig:merge_3car}
\end{figure}

Each agent admits a linear quadratic cost objective that penalizes the deviation from the reference state 
and control effort as follows
\begin{equation}
    J^i_k(x_k,u^i_k) = (x_k^i-x^i_{\textrm{ref}})^\T Q_r (x^i_k-x^i_{\textrm{ref}}) + {x^i_k}^\T Q x^i_k + u_k^i R u_k^i,
\end{equation}
where $Q_r$ penalizes the deviation from the target lateral position (lane) and deviation from the reference speed, 
and $Q$ penalizes excessive heading values. 
The reference state $x_{\textrm{ref}}^i$ specifies $i^\mathrm{th}$'s car target longitudinal velocity, 
which is set randomly at initialization, and the target lateral position, 
which is the center of the left lane for all cars. 
The collision function $h(x^i,x^j)$ is set to the Cartesian distance between two cars.
For this problem, 
the horizon is configured to 20 steps, and the initial reference control is zero.
Figure~\ref{fig:merge_3car} shows a merging maneuver with three cars.

Note that in Figure~\ref{fig:merge_3car}, the original gap between the yellow car and the purple car is not sufficient to fit the merging red car, 
so all cars have to adjust their speed for a successful merge, highlighting the ``game'' aspect of the problem.
\begin{figure}[th!pb]
  \centering
  \begin{subfigure}[b]{0.17\textwidth}
  \includegraphics[trim=290px 45px 280px 50px,clip,width=0.95\textwidth]{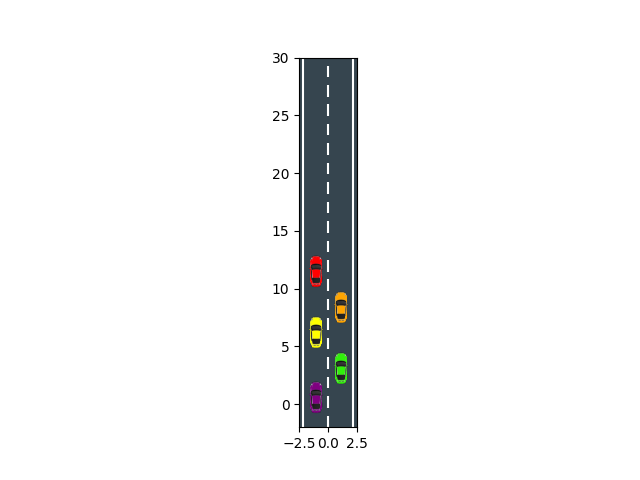}
  \caption{}
  \end{subfigure}
  \begin{subfigure}[b]{0.17\textwidth}
  \includegraphics[trim=290px 45px 280px 50px,clip,width=0.95\textwidth]{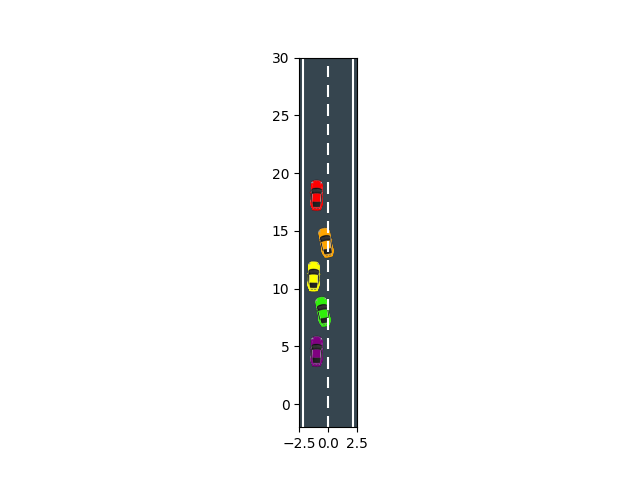}
  \caption{}
  \end{subfigure}
  \begin{subfigure}[b]{0.17\textwidth}
  \includegraphics[trim=290px 45px 280px 50px,clip,width=0.95\textwidth]{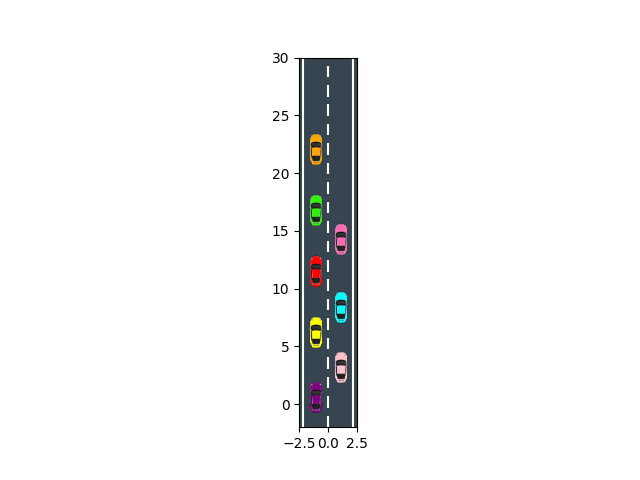}
  \caption{}
  \end{subfigure}
  \begin{subfigure}[b]{0.17\textwidth}
  \includegraphics[trim=290px 45px 280px 50px,clip,width=0.95\textwidth]{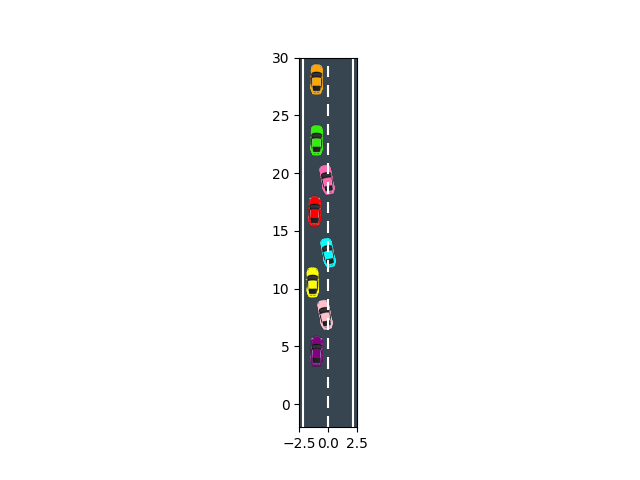}
  \caption{}
  \end{subfigure}
  \caption{Merging example with 5 and 8 cars.}
  \label{fig:merge_5car}
\end{figure}

Figure~\ref{fig:merge_5car} shows the same example with more cars, and Table~\ref{table:carmerge} shows the average solution time for different total number of cars. 
The initial states of the vehicles are randomly generated, and the algorithm terminates when the residual is smaller than $5\times 10^{-4}$ with no collision violation. 
The time shown is the average taken over 100 runs and only converged runs are included. 
For the three-car merging problem, the convergence rate is 91\% for RD3G and 37\% for iLQGame.
Table~\ref{table:carmerge} shows a clear performance advantage over the current state-of-the-art methods, 
especially when more agents are involved.
\begin{table}[h]
\centering
\caption{Average solution time (ms) of merging problems with different total car counts. 
N/C stands for ``No Convergence''.}
\begin{tabular}{|c|c|c|c|} 
 \hline
 Car Count & RD3G & ALGame\cite{algames} & iLQGame\cite{ilqgame} \\ [0.5ex] 
 \hline
 2 & 18 & 48& 324\\ 
 \hline
 3 & 29 & 95& 505\\ 
 \hline
 4 & 83 & 198 & 657\\ 
 \hline
 5 &  137 & 366 & 829\\ 
 \hline
 6 & 226 & 653 & 1037\\ 
 \hline
 7 & 339 & 1103& 2253\\ 
 \hline
 8 & 430 & 1833& N/C\\ 
 \hline
\end{tabular}
\label{table:carmerge}
\end{table}

\subsubsection{Car Racing} \label{sec:car_racing}
Our next simulation applies RD3G to an adversarial two-car racing game against a non-game-theoretic, interaction-ignorant MPC.
Both agents share the kinematic bicycle model in Section~\ref{sec:car_merge}, with the following cost function
\begin{align}
    J^i_k(x_k,u^i_k) =& Q_d (x_k^j-x_k^i) + (x_k^i-x^i_{\textrm{ref}})^\T Q_r (x^i_k-x^i_{\textrm{ref}}) \nonumber\\
    &+ {x^i_k}^\T Q x^i_k + u_k^i R u_k^i,
\end{align}
where $x_k^i$ is the state of the ego agent, and $x_k^j$ is the state of the opponent agent.
The term $Q_d (x_k^j-x_k^i)$ 
is a cost item that rewards the lead distance of the ego agent against the opponent agent. 

The agents are placed at random starting positions on a Nascar-styled track about one car length away from each other with different initial speeds. 
Each race concludes when one car has completed a full lap.
Table~\ref{table:racing} shows the win rate of RD3G against a naive MPC controller that shares an identical cost function, 
averaged over 200 simulated races with random initial states. The proposed solver shows a clear advantage over non-game-theoretic MPC.
In 72\% of the cases, the agent using RD3G successfully blocks a faster opponent from passing.

\begin{table}[h]
\centering
\begin{tabular}{|c|c|c|} 
 \hline
 Start position\textbackslash Speed & Fast & Slow \\ [0.5ex] 
 \hline
 Front & 1.0 & 0.72 \\ 
 \hline
 Rear & 0.96 & 0.16 \\ 
 \hline
\end{tabular}
\caption{Win rate of RD3G in a two-car racing game.}
\label{table:racing}
\end{table}

\subsection{Physical Experiments}

To test RD3G's real-time performance on physical robotic systems, we implemented
RD3G on a competitive racing scenario using the BuzzRacer platform~\cite{buzzracer}.

\begin{figure}[h]
    \centering
    \includegraphics[width=0.9\linewidth]{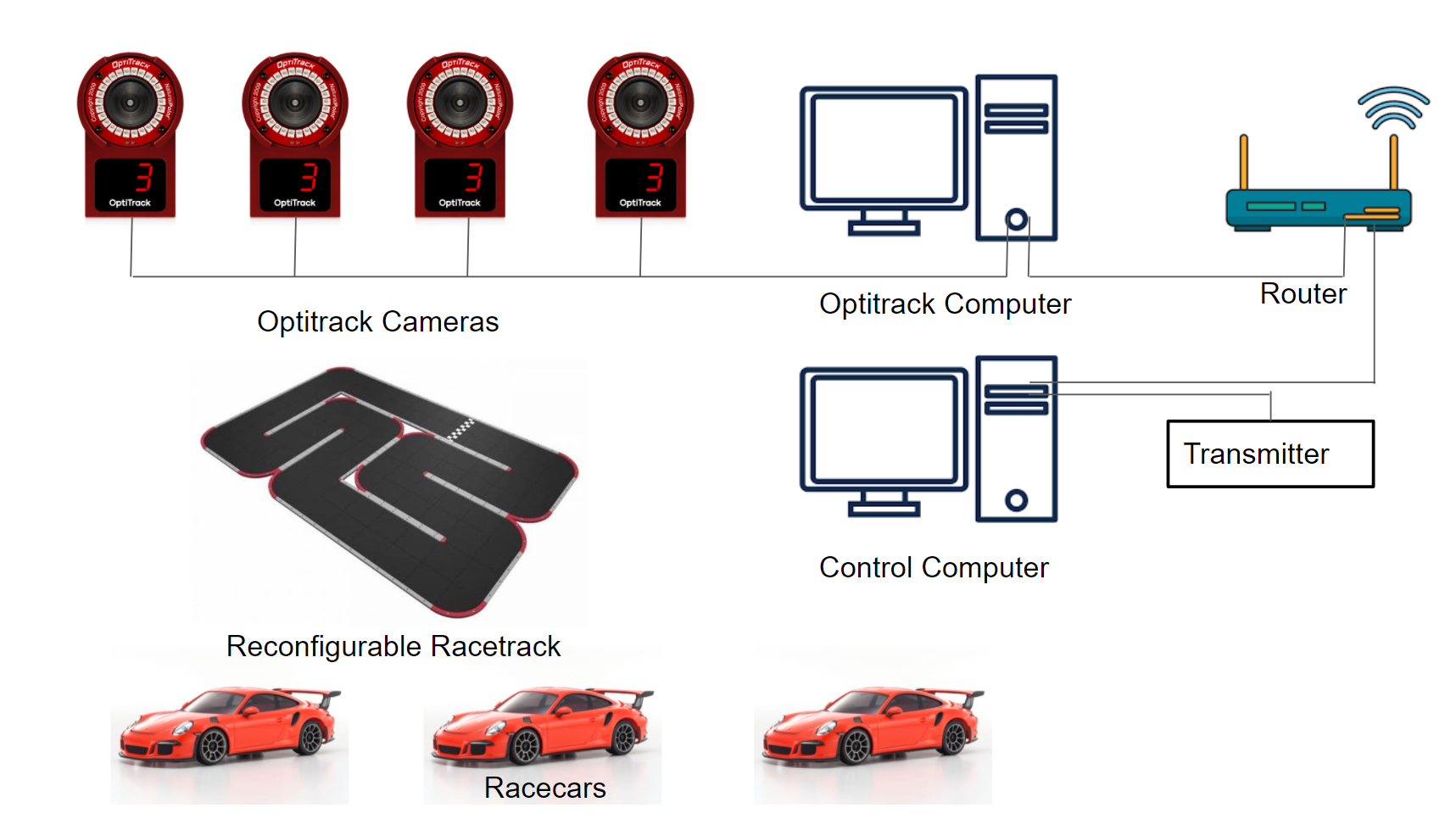}
    \caption{BuzzRacer system components.}
    \label{fig:topology}
\end{figure}

BuzzRacer is a scaled autonomous vehicle platform for validating multi-agent control algorithms. 
Figure~\ref{fig:topology} shows the system components of BuzzRacer. 
The system contains a $3\,\mathrm{m} \times 5\,\mathrm{m}$ reconfigurable race track and utilizes a visual tracking system for state measurements. 
BuzzRacer can support concurrent operations of multiple 1/28 scale cars. 
The miniature cars can reach up to 1g cornering acceleration and 3.5~m/s top speed. 

\begin{figure}[thpb]
  \centering
  \begin{subfigure}[b]{0.3\textwidth}
  \includegraphics[width=0.95\textwidth]{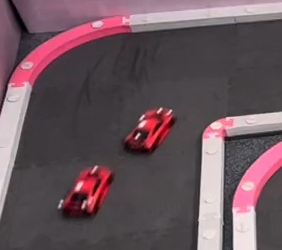}
  \caption{}
  \end{subfigure}
  \begin{subfigure}[b]{0.3\textwidth}
  \includegraphics[width=0.95\textwidth]{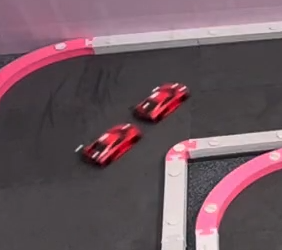}
  \caption{}
  \end{subfigure}
  \begin{subfigure}[b]{0.3\textwidth}
  \includegraphics[width=0.95\textwidth]{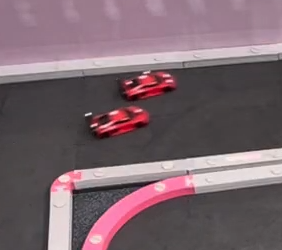}
  \caption{}
  \end{subfigure}
  \begin{subfigure}[b]{0.3\textwidth}
  \includegraphics[width=0.95\textwidth]{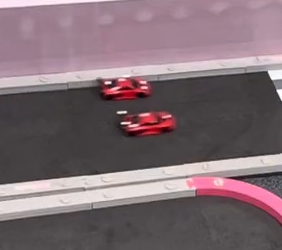}
  \caption{}
  \end{subfigure}
  \begin{subfigure}[b]{0.3\textwidth}
  \includegraphics[width=0.95\textwidth]{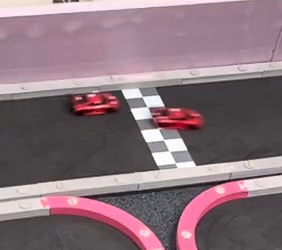}
  \caption{}
  \end{subfigure}
  \begin{subfigure}[b]{0.3\textwidth}
  \includegraphics[width=0.95\textwidth]{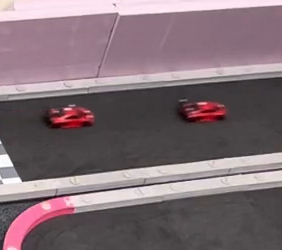}
  \caption{}
  \end{subfigure}
  \caption{Video screenshots of an overtake during the experiment.}
  \label{fig:photos}
\end{figure}

Figure~\ref{fig:photos} shows snapshots of an overtake during the experiment, 
where the vehicle controlled by RD3G overtakes the MPC-controlled car without collision. 
The average control loop frequency during the experiment is 29~Hz.
Videos of the experiments and the simulations are available on our project website: \href{https://www.github.com/Nick-Zhang1996/RD3G}{github.com/Nick-Zhang1996/RD3G}.

\section{Discussion}
\label{sec:discussion}
The main advantage of partitioning the inequality constraints in the game problem formulation is that it allows us to identify the dual variables associated with the inactive constraints, 
and therefore removing them from the residual, 
thus reducing the dimension of the optimization problem for finding the descent direction, 
which is the most time-consuming operation.

The dimension of the residual without these considerations is $N(Tn+Tm+Tn+TN) = O(TN^2)$. 
Let $c$ be the average number of agents an agent interacts with, and $k$ be the average number of interaction steps. 
We can reduce the dimensionality of the residual to $N( (Tn+kcn) + Tm + Tn + kn ) = O(TNn)$, since $k \ll T, n \ll N$.
This reduction in dimensionality has a significant influence on the solution time given that many linear solvers have a complexity of $O(c^3)$ where $c$ is the dimension of the problem~\cite{matrix_computation,numerical_la}.

\section{Conclusion}
We presented a novel solver for differential dynamic games with iterative Newton descent on primal and dual residuals.
The solver can handle collision constraints between agents. By treating active and inactive inequality constraints differently and by ignoring the dual variables associated with the inactive constraints when solving for the descent direction,
we were able to significantly reduce the size of the optimization problem. Optimizing in the state-control joint space allows for easier convergence,
but it necessitates additional steps to ensure state-control consistency.
The proposed solver performed satisfactorily on different problems and compared favorably with current state-of-art methods. 

\balance

\bibliographystyle{ieeetran}
\bibliography{refs}


\end{document}